\documentclass[twoside,11pt]{article}

\usepackage{jmlr2e}
\usepackage{listings}
\usepackage[frozencache=true,cachedir=.]{minted}
\usepackage{tcolorbox}
\usepackage{etoolbox}

\ShortHeadings{}{}
\firstpageno{1}

\begin{document}

\title{Uncertainty Quantification 360: A Holistic Toolkit for Quantifying and Communicating the Uncertainty of AI}

\author{\name Soumya Ghosh\thanks{Equal contributions. Names listed in alphabetical order.} \email ghoshso@us.ibm.com\\
\name Q. Vera Liao\footnotemark[1] \email vera.liao@ibm.com\\
\name Karthikeyan Natesan Ramamurthy\footnotemark[1] \email knatesa@us.ibm.com\\
\name Jiri Navratil\footnotemark[1] \email jiri@us.ibm.com\\
\name Prasanna Sattigeri\footnotemark[1] \email psattig@us.ibm.com\\
\name Kush R. Varshney\footnotemark[1] \email krvarshn@us.ibm.com\\
\name Yunfeng Zhang\footnotemark[1] \email zhangyun@us.ibm.com
}


\maketitle

\begin{abstract}
In this paper, we describe an open source Python toolkit named Uncertainty Quantification 360 (UQ360) for the uncertainty quantification of AI models. The goal of this toolkit is twofold: first, to provide a broad range of capabilities to streamline as well as foster the common practices of quantifying,  evaluating, improving, and communicating uncertainty in the AI application development lifecycle; second, to encourage further exploration of UQ's connections to other pillars of trustworthy AI such as fairness and transparency through the dissemination of latest research and education materials. Beyond the Python package (\url{https://github.com/IBM/UQ360}), we have developed an interactive experience (\url{http://uq360.mybluemix.net}) and guidance materials as educational tools to aid researchers and developers in producing and communicating high-quality uncertainties in an effective manner.
\end{abstract}


\section{Introduction}

Success stories of AI models are plentiful, but we have also seen prominent examples where the models behave in unexpected ways. For example, a typical failure mode of state-of-the-art prediction models is the inability to abstain from making predictions when the test data violate assumptions made during training, potentially resulting in highly confident but incorrect predictions. Hence, there is a renewed interest in improving the reliability and transparency of AI models \citep{bhatt2021uncertainty}.

A typical AI lifecycle process consists of collecting data, pre-processing it, selecting a model to learn from the data, choosing a learning algorithm to train the selected model, and performing inference using the learned model. There are inherent uncertainties associated with each of these steps. For example, data uncertainty may arise from the inability to collect or represent real-world data reliably.  Flaws in data pre-processing---whether during curation, cleaning, or labeling---also create data uncertainty. Similarly, models are only proxies for the real world and their learning and inference algorithms rely on various simplifying assumptions and thus introduce modeling and inferential uncertainties. The predictions made by an AI system are susceptible to all these sources of uncertainty. 

Reliable uncertainty quantification provides a vital diagnostic for both developers and users of an AI system. For developers, it can suggest strategies for improving the system. For example,  high data uncertainty may point towards improving the data representation process, while a high model uncertainty may suggest the need to collect more data. For users, accurate uncertainties, especially when combined with effective communication strategies, can add a critical layer of transparency and trust, crucial for better AI-assisted decision making \citep{zhang2020effect}. Trust in AI systems is a necessary condition for their successful deployment in high-stakes applications spanning medicine, finance, and the social sciences.

The research on uncertainty quantification (UQ) is long-standing but has enjoyed ever increasing interest in recent years due to the observations made above. Numerous approaches for improved UQ in AI models have been proposed. However, choosing a particular UQ method depends on many factors: the underlying model, type of machine learning task (regression vs.\ classification), characteristics of the data, and the user's goal.  If inappropriately used, a particular UQ method may produce poor uncertainty estimates and mislead users. Moreover, even a highly accurate uncertainty estimate may be misleading if poorly communicated.  To address these issues, we introduce Uncertainty Quantification 360 (UQ360) - an open-source Python toolkit. The toolkit provides a diverse set of algorithms to quantify uncertainties, metrics to measure them, methods to improve the quality of estimated uncertainties, and approaches to communicate the uncertainties effectively. In addition, we provide a taxonomy and guidance for choosing these capabilities based on the user's needs. UQ360 makes communicating UQ simple; developers can make user-centered choices by following the psychology-based guidance on communicating uncertainty estimates, ranging from concise descriptions to detailed visualizations. Altogether, UQ360's capabilities allow quantification of uncertainties to be an integral part of the AI development lifecycle.

\begin{figure}[t]
\includegraphics[width=15cm]{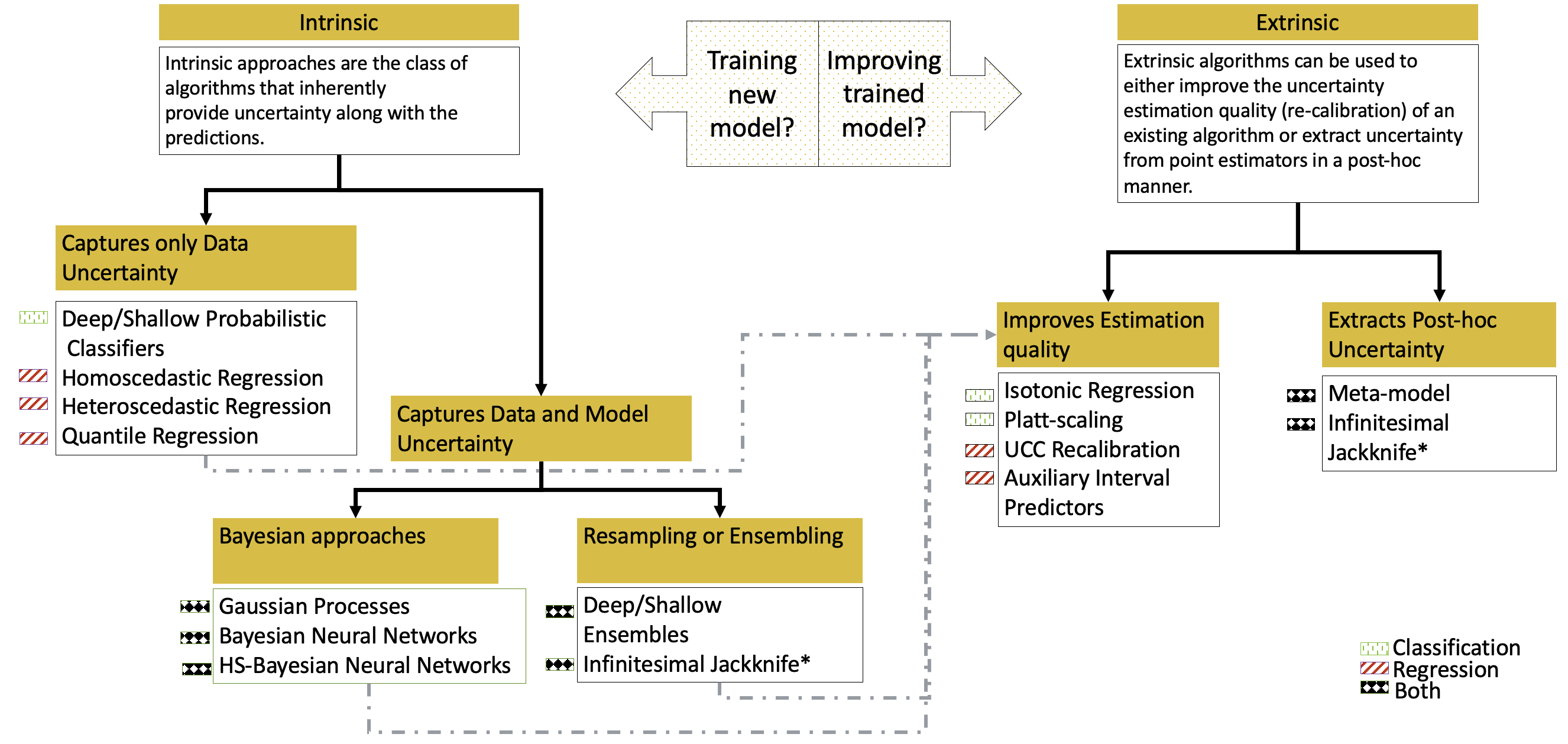}
\centering
\label{fig:workflow}
\caption{Taxonomy of the uncertainty estimation algorithms included in the UQ360 toolkit.}
\end{figure}

\begin{listing}[ht]
\begin{tcolorbox}
\begin{minted}[fontsize=\small]{python}
from sklearn.ensemble import GradientBoostingRegressor
from sklearn.datasets import make_regression
from sklearn.model_selection import train_test_split

from uq360.algorithms.blackbox_metamodel import BlackboxMetamodelRegression

X, y = make_regression(random_state=0)
X_train, X_test, y_train, y_test = train_test_split(X, y, random_state=0)
gbr_reg = GradientBoostingRegressor(random_state=0)

uq_model = BlackboxMetamodelRegression(base_model=gbr_reg)
uq_model.fit(X_train, y_train)

y_hat, y_hat_lb, y_hat_ub = uq_model.predict(X_test)
\end{minted}
\end{tcolorbox}
\caption{Use of meta-models to augment sklearn's gradient boosted regressor with prediction interval.}
\label{listing:1}
\end{listing}

\section{UQ algorithms and evaluation metrics}

UQ360 provides more than ten UQ algorithms, metrics, along with guidance to help users choose an appropriate method or metric for their use case. UQ algorithms can be broadly classified as intrinsic or extrinsic, depending on how the uncertainties are obtained from the AI models. Figure \ref{fig:workflow} lists the algorithms included in the toolkit.  \emph{Intrinsic} methods encompass approaches explicitly designed to produce uncertainty estimates along with their predictions. Amongst intrinsic methods, the toolkit includes variationally trained Bayesian neural networks (BNNs) \citep{blundell15} with Gaussian as well as heavy-tailed, sparsity-promoting Horseshoe priors \citep{ghosh2019model}, Gaussian processes \citep{gpbook}, quantile regression \citep{koenker1978regression} and neural networks with  homoscedastic and heteroscedastic noise models \citep{wakefield2013bayesian}. An infinitesimal Jackknife (IJ) based algorithm \citep{ghosh2020approximate} is also included along with these. This perturbation-based approach performs uncertainty quantification by estimating model parameters under different perturbations of the original data. Crucially, the estimation only requires the model to be trained once on the unperturbed dataset. 

The quality of estimation generated by a UQ algorithm also needs to be evaluated. Poorly calibrated uncertainties should neither be trusted nor presented to a user. UQ360 provides standard calibration metrics for classification and regression tasks to evaluate the quality of the estimated uncertainties. This includes expected calibration error (ECE) \citep{naeini2015obtaining}, Brier score ~\citep{Brier1950} for classification and prediction interval coverage probability (PICP)~\citep{chatfield1993calculating} and mean prediction interval width (MPIW) \citep{shrestha2006machine}  for regression. We also include additional diagnostic tools such as reliability diagrams~\citep {degroot1983comparison} and risk-vs-rejection rate curves \citep{el2010foundations} .  In addition, the toolbox provides a novel operation-point agnostic approaches for the assessment of prediction uncertainty estimates called the Uncertainty Characteristic Curve (UCC) \citep{navratil2021UCC}. 

\begin{listing}[ht]
\begin{tcolorbox}
\begin{minted}[fontsize=\small]{python}
from sklearn.model_selection import GridSearchCV
from uq360.utils.misc import make_sklearn_compatible_scorer
from uq360.algorithms.quantile_regression import QuantileRegression

sklearn_picp = make_sklearn_compatible_scorer(
    task_type="regression",
    metric="picp", greater_is_better=True)

base_config = {"alpha":0.95, "n_estimators":20, "max_depth": 3, 
               "learning_rate": 0.01, "min_samples_leaf": 10,
               "min_samples_split": 10}
configs  = {"config": []}
for num_estimators in [1, 2, 5, 10, 20, 30, 40, 50]:
    config = base_config.copy()
    config["n_estimators"] = num_estimators
    configs["config"].append(config)
    
uq_model = GridSearchCV(
    QuantileRegression(config=base_config), configs, scoring=sklearn_picp)
uq_model.fit(X_train, y_train)

y_hat, y_hat_lb, y_hat_ub = uq_model.predict(X_test)
\end{minted}
\end{tcolorbox}
\caption{Use of UQ360 metrics for model selection. The prediction interval coverage probability score (PICP) score is used here as the metric to select the model through cross-validation.}
\label{listing:2}
\end{listing}

For methods that do not have an inherent notion of uncertainty built into them, we use \emph{extrinsic} approaches to extract uncertainties post-hoc. The toolkit provides meta-models \citep{chen2019confidence} that generate reliable confidence measures (in classification), prediction intervals (in regression) \citep{navratil2020SRT}, and predict performance metrics such as accuracy on unseen and unlabeled data \citep{elder2020Drift}. For pre-trained models, the toolbox also provides extrinsic algorithms for potentially improving the uncertainty quality. This includes isotonic regression \citep{zadrozny2001obtaining}, Platt-scaling \citep{platt1999probabilistic},  auxiliary interval predictors \citep{thiagarajan2020building}, and UCC Recalibration \citep{navratil2021UCC}. 

\section{Implementation and communication methods}

The metrics and algorithms are designed to be scikit-learn compatible so that they can fit into developers' existing workflow.  The implementations of the algorithms and metrics are compatible with scikit-learn functions such as \textit{GridSearchCV}.  The example code block in listing \ref{listing:1} shows how to augment scikit-learn's point estimators with uncertainty using a black-box meta-model UQ algorithm. Listing \ref{listing:2} shows how to use the prediction interval coverage probability (PICP) metric to score and select models. The toolkit also comes with Jupyter notebook tutorials demonstrating the use of UQ in several industrial applications such as healthcare and finance.

UQ360 allows the choice of multiple styles of communication methods, from concise descriptions to detailed visualizations. We also provide guidance for communicating UQ to help practitioners make the choice, as informed by psychology and human-computer interaction research. For classification tasks, UQ360 provides functions to generate confidence scores. For regression tasks, UQ360 provides functions to generate the numerical ranges, visual confidence intervals, density plots, and quantile dot plots \citep{fernandes2018uncertainty}. 

The toolkit has been engineered with a common interface for all of the different UQ capabilities and is extensible to accelerate innovation by the community advancing trustworthy and responsible AI. We are open-sourcing it to help create a community of practice for researchers, data scientists, and other practitioners that need to understand or communicate the limitations of algorithmic decisions. 






\bibliography{refs}

\end{document}